\pdfoutput=1
\documentclass[11pt]{article}

\usepackage{acl}

\usepackage{times}
\usepackage{latexsym}
\usepackage{footnote}
\usepackage{hyperref}

\usepackage[T1]{fontenc}
\usepackage[utf8]{inputenc}

\usepackage{microtype}

\usepackage{covington}
\usepackage{subcaption}
\usepackage{graphicx}
\usepackage{multirow}
\usepackage{adjustbox}
\usepackage{booktabs}
\usepackage{amsmath}
\usepackage[disable]{todonotes} % use this to hide notes

\title{LSCDiscovery: A shared task on semantic change\\ discovery and detection in Spanish}

\author{Frank D. Zamora-Reina$^1$, Felipe Bravo-Marquez$^1$, Dominik Schlechtweg$^2$\\
        $^1$Department of Computer Science, University of Chile, IMFD \&  CENIA \\
        $^2$Institute for Natural Language Processing, University of Stuttgart\\
        \texttt{fzamora@dcc.uchile.cl}, \texttt{fbravo@dcc.uchile.cl},\\ \texttt{schlecdk@ims.uni-stuttgart.de}\\
}

\begin{document}
\maketitle
\begin{abstract}

We present the first shared task on semantic change discovery and detection in Spanish and create the first dataset of Spanish words manually annotated for semantic change using the DURel framework \citep{Schlechtwegetal18}. The task is divided in two phases: 1) Graded Change Discovery, and 2) Binary Change Detection. In addition to introducing a new language the main novelty with respect to the previous tasks consists in predicting and evaluating changes for all vocabulary words in the corpus. Six teams participated in phase 1 and seven teams in phase 2 of the shared task, and the best system obtained a Spearman rank correlation of $0.735$ for phase 1 and an F1 score of $0.716$ for phase 2. We describe the systems developed by the competing teams, highlighting the techniques that were particularly useful and discuss the limits of these approaches.

\end{abstract}

\section{Introduction}\label{intro}
Lexical Semantic Change Detection (LSCD) is the task of detecting words which have changed their meaning over time in a diachronic corpus of text \citep{schlechtweg-etal-2020-semeval}, usually an unsupervised task. In recent years, several LSCD shared tasks have been organized \citep{schlechtweg-etal-2020-semeval, diacrita_evalita2020,rushifteval2021}. These tasks have contributed to a better understanding of LSCD, but have also had their shortcomings: (i) they have used mainly small pre-selected sets of target words creating an unrealistic evaluation scenario for the application of computational models in historical semantics and lexicography where researchers typically aim to cover the full vocabulary of a language \citep{Kurtyigit2021discovery}, (ii) different formalizations of the LSCD task have been proposed including binary classification and ranking tasks \citep{Schlechtwegetal18,schlechtweg-walde-2020,Schlechtweg2022measurement} and these have been employed inconsistently, and (iii) none of them have focused on Spanish, despite the fact that there are more than 450 million native speakers of this language.

We tackle these shortcomings by organizing a shared task on Spanish diachronic data with a more realistic evaluation scenario requiring participants to provide Lexical Semantic Change (LSC) predictions for the full corpus vocabulary (Discovery). Additionally, we cover previous scenarios by asking participants to predict LSC only in the limited sample of annotated target words (Detection). By offering a range of additional optional tasks (defined on the same annotated data) participants are able to evaluate and compare models on various formalizations of the LSCD task. In order to derive gold LSC labels for target words, we annotate and publish the largest existing data set of semantic proximity judgments covering 100 words with approximately 62k judgments from 12 human native speakers.\footnote{The data set is available at \url{https://zenodo.org/record/6300104}.}

\section{Related Work}\label{sec:related}
The detection of lexical semantic changes is of great interest in research areas such as historical semantics, lexicography, linguistics and NLP. For a comprehensive review of the literature on the area we refer the reader to the recent surveys \citep{2018arXiv181106278T,kutuzov-etal-2018-diachronic,hengchen2021challenges}. In previous years several shared tasks have been organized: SemEval-2020 Task 1  \citep{schlechtweg-etal-2020-semeval} for English, German, Latin, and Swedish, DIACR-Ita for Italian \citep{diacrita_evalita2020}, and RuShiftEval for Russian \citep{rushifteval2021}.\footnote{There was also a student shared task on German data \citep{AhmadEtal2020}.} All shared tasks applied an evaluation setup where LSC was measured between pairs of time periods.%\footnote{For RuShiftEval, there are multiple pairs of time periods.} 

\paragraph{SemEval} used a total of 156 target words for all languages with no development/test split. Approximately half of these were drawn from etymological dictionaries or research literature, while the other half was drawn from the corpus vocabularies by selecting lemmas with similar POS and frequency as the first half of target words. Target word occurrences in sentences (usages) were combined into pairs and these were annotated for their semantic proximity \citep{Schlechtweg2021dwug}. Target words were excluded if they had a high number of undecidable use pairs or were annotated too sparsely. Sense clusters were inferred from the annotation. From the clusters a binary (sense loss/gain vs. none) and a graded (Jensen-Shannon distance between cluster distributions) change score were derived and used to evaluate participants on a corresponding binary classification and ranking task.
% The annotated data was represented in a graph and clustered. 

\paragraph{DIACR-Ita} used a total of 18 target words with no development/test split. All of these were drawn from an etymological dictionary. Target word usages were annotated with word sense definitions. Words with a high number of OCR errors and annotator disagreements were excluded. From the annotation Binary Change scores similar to SemEval were derived and used to evaluate participants on a binary classification task.

\paragraph{RuShiftEval} used a total of 111 target words (all nouns) split into 12 for development and 99 for testing. These were selected in a similar procedure to SemEval: approximately half of these were drawn from etymological dictionaries, research literature or ``invented'' by the authors, while the other half was drawn from the corpus vocabularies by selecting lemmas with similar POS and frequency as the first half of target words. Target word usages from different time periods were combined into usage pairs and annotated for semantic proximity. From these the DURel COMPARE score (see Subsection \ref{sub:optional_tasks} for more details) \citep{Schlechtwegetal18} was derived, which can be seen as an approximation of SemEval's Graded Change score \citep{Schlechtweg2022measurement}. Participants were evaluated in a ranking task on the COMPARE scores.\\  

\noindent As we can see, target words in previous shared tasks have been strongly preselected and systems have been evaluated on different tasks. They have also yielded (seemingly) contradictory results: while type-based model architectures have dominated in SemEval and DIACR-Ita, token-based architectures have dominated in RuShiftEval. In all tasks clustering-based models have shown rather low performance.

\section{Task description}\label{sec:task}

Our task was designed in two phases:
\begin{enumerate}
    \item Graded Change Discovery, and
    \item Binary Change Detection.
\end{enumerate}
Note that \textit{discovery} introduces additional difficulties for models as compared to the more simple semantic change \textit{detection}, e.g. because a large number of predictions is required and the target words are not preselected, balanced or cleaned \citep[cf.][]{Kurtyigit2021discovery}. Yet, discovery is an important task, with applications such as lexicography where dictionary makers aim to cover the full vocabulary of a language.

\subsection{Phase 1: Graded Change Discovery}\label{sub:discovery}

Similar to \citet{Kurtyigit2021discovery}, we define the task of \textbf{Graded Change Discovery} as follows:
\begin{itemize}
  \item[] Given a diachronic corpus pair $C_1$ and $C_2$, rank the intersection of their (content-word) vocabularies according to their degree of change between $C_1$ and $C_2$.
\end{itemize}
The participants were asked to rank the set of content words in the lemma vocabulary intersection of $C_1$ and $C_2$ according to their degree of semantic change between $C_1$ and $C_2$ where a higher rank means stronger change. The true degree of semantic change of a target word $w$ was given by the Jensen-Shannon distance \citep{Lin91divergencemeasures,DonosoS17} between $w$'s word sense frequency distributions in $C_1$ and $C_2$ \citep[cf.][]{schlechtweg-etal-2020-semeval}. The two word sense frequency distributions were estimated via human annotation of word usage samples for $w$ from $C_1$ and $C_2$ (see Subsection \ref{sub:annotation}).  Participants' predictions were \textit{not} evaluated on the full set of target words, as this would be unfeasible to annotate, but on an (unpublished) random sample of words from the full set of target words. The predictions were scored against the ground truth via Spearman's rank-order correlation coefficient \citep{bolboaca2006pearson}.

\subsection{Phase 2: Binary Change Detection}\label{sub:binary_change_detection}

Similar to \citet{schlechtweg-etal-2020-semeval}, we define the task of \textbf{Binary Change Detection} as follows:
\begin{itemize}
  \item[] Given a target word $w$ and two sets of its usages $U_1$ and $U_2$, decide whether $w$ lost or gained senses from $U_1$ to $U_2$, or not.
\end{itemize}
The participants were asked to classify a pre-selected set of content
words into two classes, 0 for no change and 1 for change. The
true binary labels of word \emph{w} were inferred from \emph{w}'s
word sense frequency distributions in $C1$ and $C2$ (see Subsection \ref{sub:discovery}). Participants' predictions were scored against
the ground truth with the following metrics: F1 (main metric), Precision, and Recall.
A crucial difference compared to Graded Change Discovery was that the
public target words corresponded exactly to the hidden words on which we
evaluated. Also, we published the usages sampled for annotation. Hence, participants could work with the exact annotated data, which was not possible in the first phase where participants could only work with the full corpora (from which the usages for annotation were sampled).

\subsection{Optional tasks}\label{sub:optional_tasks}
Participants could submit predictions for several optional tasks: 

\paragraph{Graded Change Detection} was defined similar to Graded
Discovery. The only difference was that the public target words corresponded
exactly to the hidden words on which we evaluated. 
Participants were scored with Spearman correlation.

\paragraph{Sense Gain Detection} was similar to Binary Change Detection. However, only words which gained (not lost) senses receive label 1. Participants were scored with F1, Precision and Recall.

\paragraph{Sense Loss Detection} was similar to Binary Change Detection. However, only words which lost (not gained) senses received label 1. Participants were scored with F1, Precision and Recall.

\paragraph{COMPARE} asked participants to predict the negated DURel
COMPARE metric \cite{Schlechtwegetal18}. This metric is defined as the
average of human semantic proximity judgments of usage pairs for \emph{w}
between $C1$ and $C2$.\footnote{Contrary to the original metric we first
  take the median of all annotator judgments for each usage pair and
  then average these values. For details see: 
  \url{https://github.com/Garrafao/WUGs}.} It can be seen as an approximation of JSD (Graded Change) \citep{Schlechtweg2022measurement}. Participants were scored with Spearman correlation.\\

\noindent Participants' submission files only needed to include predictions corresponding to the obligatory tasks in order to get a valid submission. They did not see the leaderboard while the evaluation phases were running. Furthermore, participants only had three valid submissions for each evaluation phase.\footnote{We decided not to include the binary subtasks in phase 1, as the usage samples were not published which meant that participants needed to work with the full corpora instead of the samples on which the gold scores were inferred. We assumed that the sampling error between usages in the full corpora and our samples is much larger for Binary Change than for Graded Change \citep[cf.][]{Schlechtweg2022measurement}.}

\section{Data}\label{sec:data}
In this section, we describe the corpora, the selection process of target words, the sampling of usages and their annotation. Moreover, we explain how the target words were presented to the participants considering the two phases of the shared task.

\subsection{Corpora}
We created two corpora covering disjoint time periods: 1810 to 1906 (old corpus, $C1$) and 1994 to 2020 (modern corpus, $C2$) (see Table \ref{tab:corpora}). The former was created using different sources freely available from Project Gutenberg\footnote{\url{https://www.gutenberg.org/browse/languages/es}} and the latter using different sources available from the OPUS project\footnote{\url{https://opus.nlpl.eu/}} \citep{TIEDEMANN12.463}. For the old corpus, all the sources collected were concatenated. As for the modern corpus, four datasets were used: Spanish portion of TED2013, Spanish portion of News-Commentary v16, Spanish portion of MultiUN and Spanish version of Europarl corpus. TED2013 was used in its entirety, while $50$ snippets with $5000$ lines each were extracted from the other datasets by cutting the corpora into snippets of the mentioned size and randomly choosing $50$ of them. 

\begin{table}[t]
\centering
\begin{tabular}{lll}
\hline
Corpus & Time period & Tokens \\ \hline
Old corpus ($C_1$) & 1810--1906 & $\sim 13M$ \\
Modern corpus ($C_2$) & 1994--2020 & $\sim 22M$
\end{tabular}
\caption{Sizes of both corpora.}
\label{tab:corpora}
\end{table}

Both corpora were parsed using spaCy \citep{spacy}.\footnote{Find details issues in Appendix \ref{sec:lemma}.} Each corpus contains four versions of the original dataset (raw, tokenized, lemmatized and POS-tagged). 
\todo[inline]{Frank, please put the link to the extracted corpora.}

\hypertarget{target-words}{%
\subsection{Target words}\label{target-words}}
\subsubsection{Phase 1 (Graded Discovery)}
\paragraph{Public target words} was a list of 4385 words created in the following way: we first took the corpus vocabulary intersection from the lemmatized versions of both corpora. Then we removed words below a minimum frequency threshold of 40 for the old corpus and 73 for the modern corpus.\footnote{40 was chosen by us for the old corpus and then we calculated 73 for the new corpus to reflect the same proportion of the frequency threshold to corpus size.} Then we removed all non-content words, i.e., we left only nouns, verbs, adjectives and adverbs. The final list of target words was published and participants were required to submit results for all 4385 words in the development and evaluation phase 1.

\paragraph{Hidden target words}
The large number of public target words was crucial to our task. However, it was not feasible to annotate all of them. Hence, we only annotated a subset of the public target words for semantic change. Participants' predictions for development and evaluation phase 1 were evaluated only on this subset of target words, which remained hidden from the participants. We selected the hidden target words in the following way: Initially, a list of $15$ changing words was selected by scanning etymological dictionaries and consulting with a linguistic specialist to obtain words for changes from $C1$ to $C2$. Likewise, it was verified that these words were in both corpora. Additionally, a list of $85$ words were randomly sampled from the public target words. The $85 + 15 = 100$ words were annotated as described in Section \ref{sub:annotation}. Then, 20 words were excluded based on inter-annotator agreement.\footnote{We removed target words with agreements of less than $0.3$ Krippendorf's $\alpha$ and less than $0.3$ on a version of Krippendorf's $\alpha$ where expected disagreements were calculated from the full annotated data (instead of for each word separately). The latter measure is less sensitive to skewed judgment distributions for individual words.}
The remaining set of 80 target words were split randomly into two groups, 20 words for the development set and 60 for the evaluation set (see Table \ref{tab:data}). Uploaded submissions were scored against these 20/60 annotated words during development/evaluation phases.

\subsubsection{Phase 2 (Binary Detection)}

The target words corresponded to the 20/60 hidden words from Phase 1 for development/evaluation. There it was no distinction here between public and hidden target words. Participants also got access to the annotated usages (20+20 from each corpus). Uploaded submissions were scored against the 20/60 public annotated words.

\subsection{Word usages}
All occurrences of the target words per corpus were extracted according to the lemma. Then, 20 usages were randomly sampled per target word from each corpus.

\begin{table}[t]
\centering
\tabcolsep=0.11cm
\begin{tabular}{ll}
\multirow{4}{*}{$\Bigg\uparrow$} &4: Identical\\
 &3: Closely Related\\
 &2: Distantly Related\\
 &1: Unrelated\\
\end{tabular}
\caption{DURel relatedness scale \citep{Schlechtwegetal18}.}\label{tab:scales}
%\vspace{-8ex} 
\end{table}

\begin{figure*}[t]
    \begin{subfigure}{0.33\textwidth}
\frame {        \includegraphics[width=\linewidth]{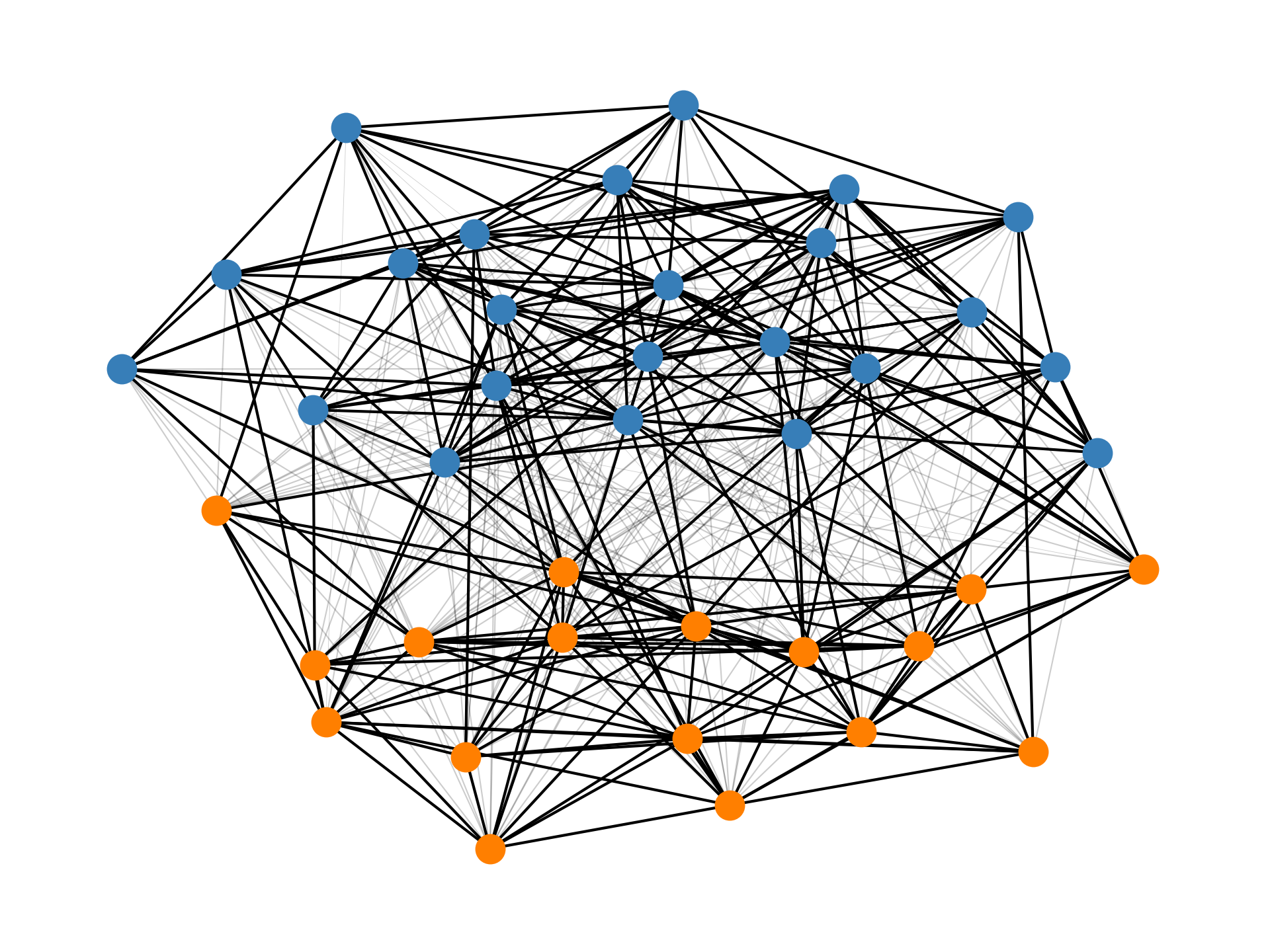}}
        \caption*{$G$, $D=(23,17)$}
    \end{subfigure}
    \begin{subfigure}{0.33\textwidth}
\frame{        \includegraphics[width=\linewidth]{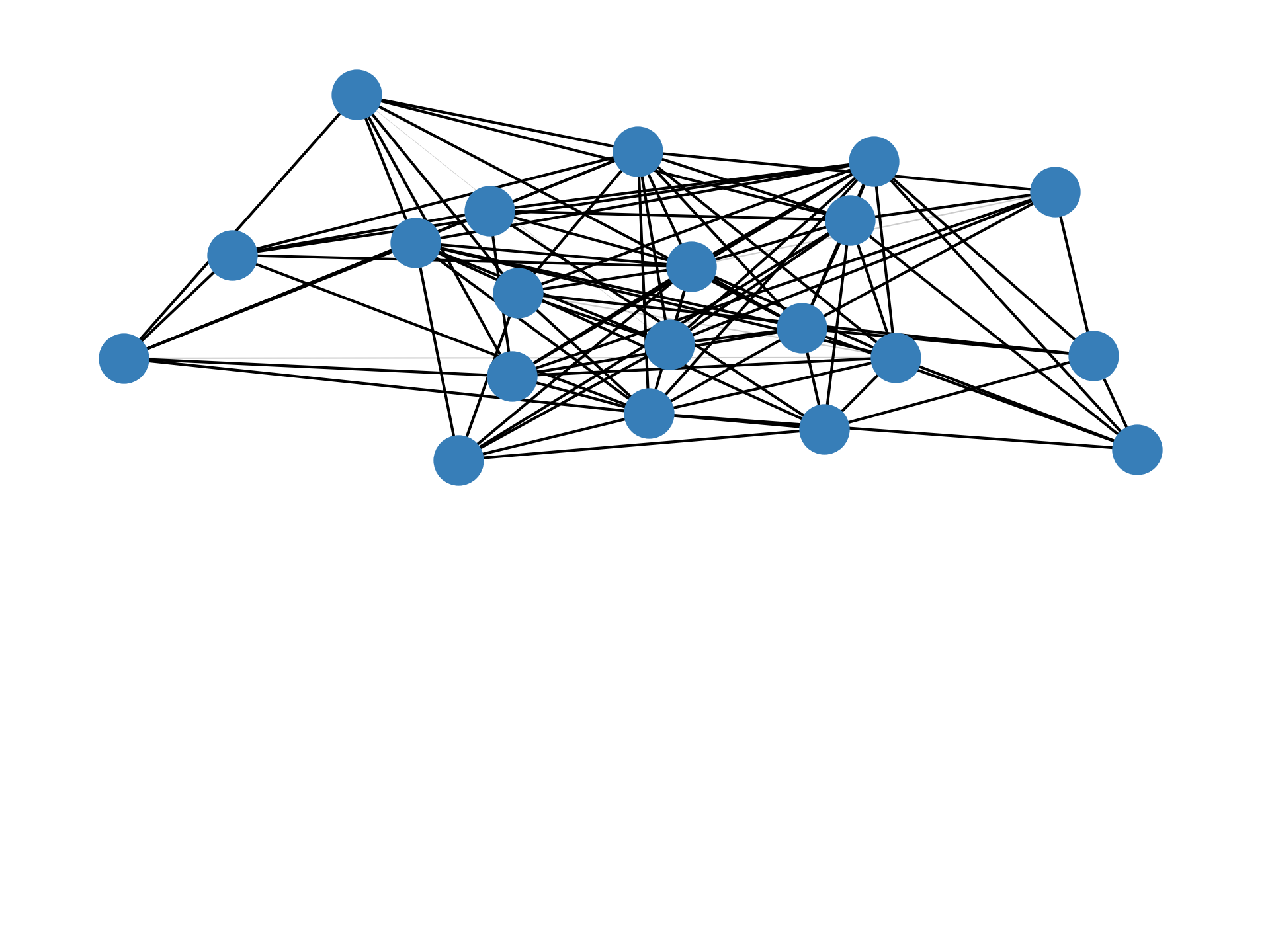}}
        \caption*{$G_1$, $D_1=(20,0)$}
    \end{subfigure}
    \begin{subfigure}{0.33\textwidth}
\frame{        \includegraphics[width=\linewidth]{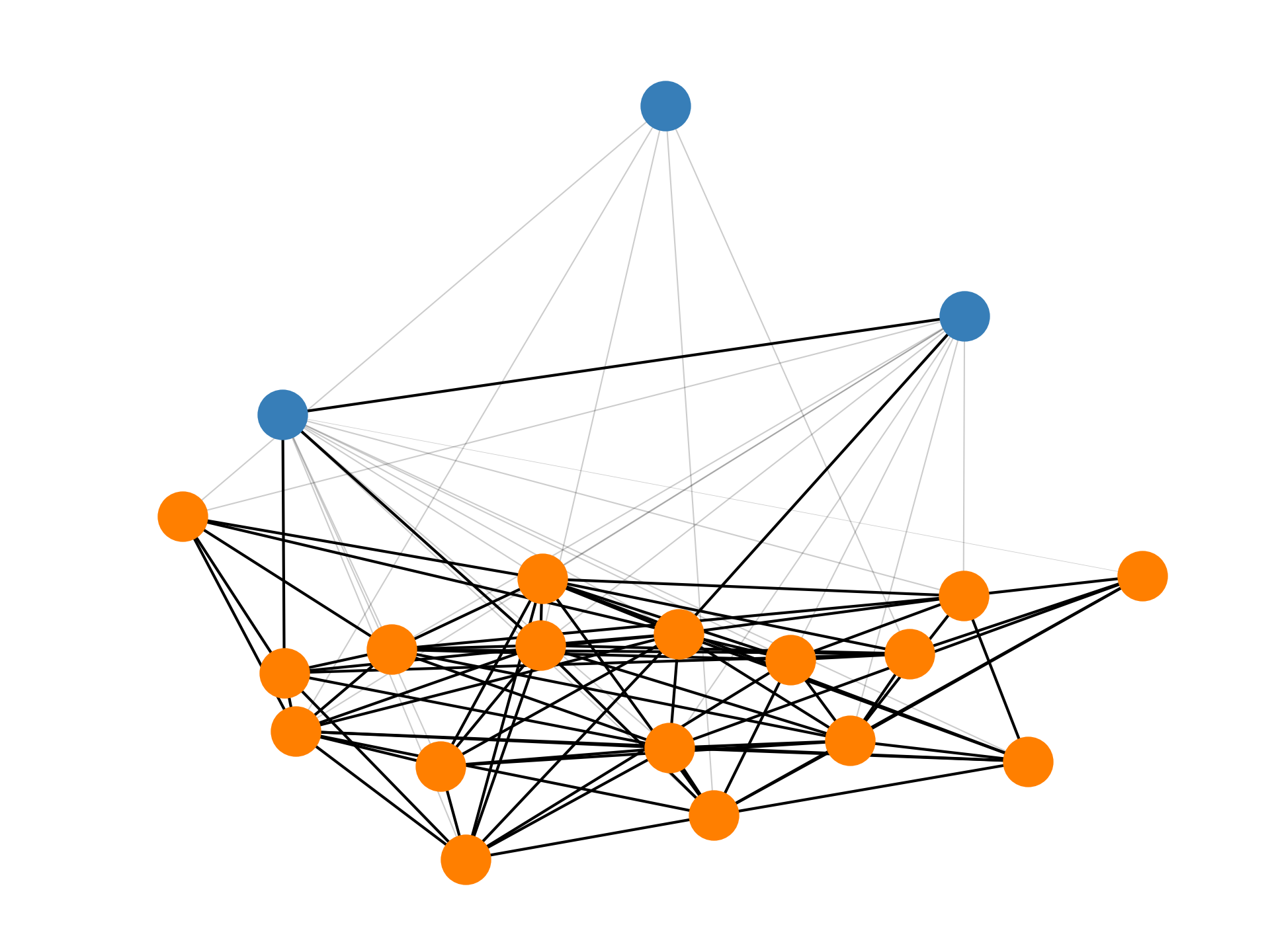}}
        \caption*{$G_2$, $D_1=(3,17)$}
    \end{subfigure}
    \caption{Word Usage Graph \textit{servidor} (left), subgraphs for old corpus $G_1$ (middle) and for modern corpus $G_2$ (right). The colors correspond to the clusters. \textbf{black}/\textcolor{gray}{gray} lines indicate \textbf{high}/\textcolor{gray}{low} edge weights.} \label{fig:servidor}
\end{figure*}

\subsection{Annotation}\label{sub:annotation}

We applied the SemEval procedure to annotate target word usages, as described in \citet{schlechtweg-etal-2020-semeval,Schlechtweg2021dwug}. Annotators were asked to judge the semantic relatedness of pairs of word usages, such as the two usages of \textit{servidor} in (\ref{ex:1}) and (\ref{ex:2}), on the scale in Table \ref{tab:scales}. 
\begin{example}\label{ex:1}
Todo esto lo hago con mi iPhone; se va derecho al \textbf{servidor}, allí se hace el trabajo de archivo, clasificación y ensamble.\\
`\em I do all this with my iPhone; it goes straight to the \textbf{server}, there the work of archiving, sorting and assembling is done.'
\end{example}%
\begin{example}\label{ex:2}
Llamó a grandes voces a sus \textbf{servidores}, y únicamente le contestó el eco en aquellas inmensas soledades, y se arrancó los cabellos y se mesó las barbas, presa de la más espantosa desesperación.\\
`\em He called out to his \textbf{servants}, and only the echo in those immense solitudes answered him, and he pulled out his hair and ruffled his beard, prey to the most frightening desperation.'
\end{example}
The annotated data of a word was represented in a Word Usage Graph (WUG), where vertices represented word usages, and weights on edges represented the (median) semantic relatedness judgment of a pair of usages such as (\ref{ex:1}) and (\ref{ex:2}). The final WUGs were clustered with correlation clustering \citep{Bansal04,schlechtweg-etal-2020-semeval,Schlechtweg2021dwug} (see Figure \ref{fig:servidor}, left) and split into two subgraphs $G_1$ and $G_2$ representing nodes from subcorpora $C_1$ and $C_2$ respectively (middle and right). Clusters were then interpreted as word senses and changes in clusters over time as lexical semantic change.\footnote{We used \citet{schlechtweg-etal-2020-semeval,Schlechtweg2021dwug}'s code provided at \url{https://www.ims.uni-stuttgart.de/data/wugs}.}

In contrast to \citeauthor{schlechtweg-etal-2020-semeval}, we used the openly available DURel interface for annotation and visualization.\footnote{\url{https://www.ims.uni-stuttgart.de/data/durel-tool}.} This also implied a change in sampling procedure, as the system implemented only random sampling of usage pairs (without SemEval-style optimization, i.e., sampling in rounds with connection of clusters). For each target word we sampled $|U_1|=|U_2|=20$ usages (sentences) per subcorpus ($C_1$, $C_2$) and uploaded these to the DURel system, which presented usage pairs to annotators in randomized order. We recruited twelve Spanish native speakers (4 Chileans, 4 Colombians, 2 Cubans, 1 Spaniard and 1 Venezuelan). All had university level education, while seven had a background in linguistics of which two had one in historical linguistics. We monitored agreement between annotators during the annotation process and discussed some strong annotation disagreements with certain annotators. This led to the exclusion of one annotator early in the process who often completely inverted the annotation scale (e.g. judged 1 while agreeing that the two usages have identical meanings).
  
Similar to \citet{schlechtweg-etal-2020-semeval}, we ensured the robustness of the obtained clusterings by continuing the annotation of a target word until all clusters in its WUG were connected by at least one judgment.\footnote{Note that this condition was more strict than \citet{schlechtweg-etal-2020-semeval}'s where only connection of multi-clusters (clusters with more than one usage) was guaranteed. Their condition was always met in our data.} For 16 words the annotation had to be stopped before this condition was met. We manually inspected the unconnected clusters of some words and concluded that missing connections did not lead to clustering errors. 

We finally labeled a target word as Binary Change if it gained or lost a cluster over time. For instance, \textit{servidor} in Figure \ref{fig:servidor} was labeled as change as it gained the orange cluster from $C_1$ to $C_2$. Consequently, \textit{servidor} was also labeled as gaining a sense; but not as losing a sense, since the blue cluster persists. Graded Change was defined as the Jensen-Shannon distance between the normalized cluster frequency distributions $D_1$ and $D_2$ yielding a high value of $0.82$ (ranges between $0.0$ and $1.0$) for \textit{servidor}, as sense probabilities changed drastically. The negated COMPARE score was derived by averaging over all graph edges with nodes from different time periods and negating this value, yielding a high score of $-1.97$ (ranges between $-4.0$ and $-1.0$) for \textit{servidor}.\footnote{Find a more detailed discussion of different change scores in \citet{schlechtweg-etal-2020-semeval} and \citet{Schlechtweg2022measurement}.} Following \citet{schlechtweg-etal-2020-semeval} we used $k$ and $n$ as lower frequency thresholds for the binary notions to avoid that small random fluctuations in sense frequencies caused by sampling variability or annotation error were misclassified as change. As proposed in \citet{Schlechtweg2021wugs} for comparability across sample sizes we set $k = 1 \leq 0.01*|U_i| \leq 3$ and $n = 3 \leq 0.1*|U_i| \leq 5$, where $|U_i|$ was the number of usages from the respective time period.\footnote{That is, $k$ was always between $1$ and $3$. There are three possible cases: $k=1$ if $0.01*|U_i| \leq 1$, $k=0.01*|U_i|$ if $1 < 0.01*|U_i| < 3$, $k=3$ if $0.01*|U_i| \geq 3$. Similarly for $n$.} This resulted in $k=1$ and $n=3$ for all target words.

Find an overview over the final set of WUGs in Table \ref{tab:data}. We reached an inter-annotator agreement of Krippendorff's $\alpha=.53$ and Spearman's $\rho=.57$ which was comparable to previous studies \citep[e.g.][]{Schlechtwegetal18,rodina2020rusemshift,Kurtyigit2021discovery,Baldissin2021diawug}.\footnote{We provide WUGs as Python NetworkX graphs, descriptive statistics, inferred clusterings, change values and interactive visualizations for all target words and the respective code at \url{https://www.ims.uni-stuttgart.de/data/wugs} (DWUG ES).}

\begin{table*}[t]
\centering
\tabcolsep=.09cm
\begin{tabular}{ c | c c c c c c c c c c c c}            
\toprule                           
\textbf{Data set}  &  $\mathbf{n}$  &  \textbf{N/V/A} &  $\mathbf{|U|}$  & \textbf{AN} & \textbf{ JUD } &  \textbf{AV }  & \textbf{KRI } & \textbf{SPR }  &  \textbf{UNC } & \textbf{ LOSS } & \textbf{LSC$_B$}& \textbf{LSC$_G$} \\
\midrule                           
development & 20 & 13/4/3 & 40 & 10 & 12k & 40 & .53 & .59 & 0 & .53 & .55 & .39\\
evaluation & 60 & 30/14/16 & 40 & 12 & 38k & 40 & .58 & .60 & 0 & .45 & .47 & .37 \\
discarded & 20 & 8/6/6 & 40 & 12 & 12k & 40 & .27 & .33 & 0 & .52 & .30 & .18 \\
\midrule                           
full & 100 & 51/24/25 & 40 & 12 & 62k & 40 & .53 & .57 & 0 & .48 & .45 & .34 \\
\bottomrule                           
\end{tabular}
\caption{Overview target words. $n$ = no. of target words, N/V/A = no. of nouns/verbs/adjectives+adverbs, $|U|$ = avg. no. of usages per word, AN = no. of annotators, JUD = total no. of judged usage pairs, AV = avg. no. of judgments per usage pair, KRI = Krippendorff's $\alpha$, SPR = weighted mean of pairwise Spearman, UNC = avg. no. of uncompared multi-cluster combinations, LOSS = avg. of normalized clustering loss * 10, LSC$_{B/G}$ = mean binary/Graded Change score.}
\label{tab:data}
\end{table*}

\section{Systems}\label{sec:systems}

We now summarize the baseline systems as well as the systems and resources used by the participating teams.

\subsection{Baselines}\label{baselines}
For both phases we use five baselines:

\paragraph{baseline1} \textit{Skip-Gram with Negative Sampling + Orthogonal Procrustes + Cosine Distance (SGNS+OP+CD)}
This approach learned vector representations for each word (type-based) in two input corpora with a shallow neural language model \citep{Mikolov13a,Mikolov13b}.\footnote{As parameters we chose dim=$100$, window size=$10$, epochs=$5$, number of negative samples=$5$, subsampling threshold=$0.001$ \citep[cf.][]{kaiser-etal-2020-IMS}.} These were then aligned using Orthogonal Procrustes \citep{ Hamilton16}. For phase 1, the method computed Graded Change as the cosine distance between old and modern vectors for all words in the vocabulary. This same value was used in the COMPARE subtask. In phase 2, binary predictions were computed by setting a threshold to the cosine distances, which was calculated as the sum between the mean and the standard deviation (std) of all these distances \citep{kaiser-etal-2020-roots}.  All words with values above the threshold were classified as \textit{change}, and values below were classified as \textit{no change}. This approach has shown high performance in several previous studies and shared tasks \citep{Schlechtwegetal19, pomsl-lyapin-2020-circe, kaiser-etal-2020-roots, prazak-etal-2020-uwb}.

 \paragraph{baseline2  } \textit{Normalized Log-Transformed Frequency Difference (FD)}
 For phase 1, this method calculated the frequency of each target word in each of the two corpora, normalized it by the logarithm of the total corpus frequency and then calculated absolute differences between these values as a measure of change. We submitted these values for the change graded and COMPARE subtasks. For phase 2, the method applied the same thresholding approach used in baseline1. For the sense loss subtask, it first verified that the target word presents change using the value of the change binary subtask. Then, if the differences were negative, the words were classified as loss $=1$ and as loss $=0$ otherwise. For sense gain the labeling is reversed.

\paragraph{baseline3}
 \textit{Grammatical profiles} were generated from tagged and parsed corpora \citep{kutuzov-etal-2021-grammatical}. These profiles were essentially frequency vectors of various morphological and syntactic features (for example, \textit{case = Nominative}, or \textit{syntax role = subject}) for a given word in a given historical corpus.
The cosine distance between the profile vectors of the same word for the two periods was used as an estimate of graded semantic change and COMPARE. Binary predictions were generated from ordered lists of graded scores for all target words by applying an offline change-point detection algorithm based on dynamic programming. The algorithm finds a point (a word) in an ordered list of scores, where the scores become significantly higher. This word and all words with score values above it were assigned the class ``changed''. This baseline did not produce predictions for the sense loss and sense gain subtasks.\footnote{All results for baseline3 were computed and submitted by Andrey Kutuzov using the code at \url{https://github.com/glnmario/semchange-profiling}.}

\paragraph{baseline4 } \textit{Minority class} This baseline produced predictions by labeling each word with the minority class label of the respective Binary Change score (change binary, loss, gain). This is label $1$ (change) in all cases. It only applied to phase 2.

\paragraph{baseline5 } \textit{Random baseline} This baselines produced random predictions for all subtasks in both phases. For phase 1, we generated random values between $0$ and $1$ from a uniform distribution for all hidden target words and computed Spearman correlation with the gold scores. This process was repeated $100$ times and we reported the average performance over all repetitions. For phase 2, we used a parallel procedure generating random labels $\in \{0,1\}$ from a uniform distribution.\footnote{Baseline3, baseline4 and baseline5 were added after the shared task finished.}

\subsection{Participating systems}\label{sec:partsystems}

Below we present a summary of the methods developed by the participants:\footnote{The descriptions are based on the system description papers submitted by the participating teams, with the exception of Rombek who did not provide a paper but gave us a brief description by e-mail.}

\paragraph{HSE} \textit{\citep{hse-lcsdiscovery}}
This team participated with two different methods.
The first consisted of fine-tuning BERT \citep{devlin-etal-2019-bert} on the lemmatized versions of the corpora in order to extract embeddings of the target words separately for each period, which are then clustered using K-means. Graded Change was estimated as the average cosine distance between all pairs of cluster centroids in the first and second periods. In order to estimate Binary Change, the Graded Change scores were thresholded by clustering them into two clusters.

The second method was based on grammatical profiles \cite{kutuzov-etal-2021-grammatical}. The frequency of morphological and syntactic categories for each target word in both corpora \citep[parsed with UdPipe, ][]{straka-strakova-2017-tokenizing} were counted and used as features in two time-specific vectors. Graded Change was measured by the cosine distance between these vectors, while Binary Change was measured by thresholding the graded scores.

\paragraph{GlossReader} \textit{\citep{glossreader-lcsdiscovery}}
This system fine-tuned the XLM-R multilingual language model \citep{conneau2019unsupervised} as part of a gloss-based Word Sense Disambiguation (WSD) system on a large English WSD dataset. It employed zero-shot cross-lingual transferability to build contextualized embeddings for Spanish data. The Graded Change score for each word was calculated as the Average Pairwise (Manhattan) Distance (APD) between the embeddings for (non-preprocessed) word usages in the old and new corpus. Binary changes were estimated by thresholding these scores. For the sense gain and sense loss subtasks the same predictions were reused.

\paragraph{UAlberta} \textit{\citep{ualberta-lcsdiscovery}}
This team applied different methods to the two subtasks.
For Graded Change Discovery, they followed the design of CIRCE \citep{pomsl-lyapin-2020-circe} and computed distances based on both static (type-based) and contextual (token-based) embeddings, with their relative weights tuned on the development set. For static embeddings, they used SGNS+OP+Euclidean Distance on the lemmatized versions of the corpora. For contextual embeddings, the XLM-R model was trained on the combined corpus (tokenized) to predict masked instances of the target words and Graded Change was measured using Euclidean APD.
For Binary Change Detection, they framed the task as a WSD problem, creating sense frequency distributions for target words in the old and modern corpus with an end-to-end WSD system \citep{orlando-etal-2021-amuse}. It was assumed that the word semantics has changed if: (1) a sense is observed in the modern corpus but not in the old corpus (or vice versa), or (2) the relative change for any sense exceeds a tuned threshold.

\paragraph{CoToHiLi} \textit{\citep{cotohili-lcsdiscovery}}
This team proposed a type-based embedding model combined with hand-crafted linguistic features. The system computed several features for every target word based on embedding distances between time periods and linguistic hand-crafted features, which were then weighted into an ensemble model to predict the final score. First, the system obtained word embeddings separately on the two corpora (tokenized) with the Continuos Bag-of-Words (CBOW) model \citep{Mikolov13a, Mikolov13b}, which were then aligned to obtain a common embedding space. The alignment algorithms used were: supervised alignment using a seed word dictionary and a linear mapping method, a semi-supervised algorithm and unsupervised alignment based on adversarial training \citep{artetxe2016,artetxe2017acl,artetxe2018aaai, artetxe-etal-2018-uncovering}. Finally, cosine distance between embeddings of the same word in different corpora was used as an indicator of graded semantic change. For the binary task, the system used thresholding the graded scores.

\paragraph{DeepMistake} \textit{\citep{deepmistake-lcsdiscovery}}
This team employed a Word-in-Context (WiC) model, i.e., a model designed to determine if a particular word has the same meaning in two given contexts. In essence, they attempted to directly apply a model trained on a related task to our problem. The WiC model was initially trained by fine-tuning the XLM-R language model on the Multilingual and Cross-lingual Word-in-Context (MLC-WiC) dataset \citep{martelli-etal-2021-semeval}. Subsequently, it was further fine-tuned on the provided annotations for the development set in this shared task and on the Spanish portion of the multi-language XL-WSD  dataset \citep{pasini2021xl}. Graded Change was measured similarly to APD by averaging same-sense probabilities between embeddings for usages (no preprocessing) from different time periods. For the change binary subtask, the authors applied thresholding to the Graded Change scores, for the sense gain and sense loss subtasks the same predictions were reused.

They also experimented with clustering by representing word usages and their same-sense probabilities in a weighted undirected graph, which was then clustered with Correlation Clustering. Graded Change was measured with JSD, while Binary Change was measured with the Binary Change score definition from Section \ref{sub:annotation}.

\paragraph{BOS} \textit{\citep{bos-lcsdiscovery}}
The system described by this team was based on generating lexical substitutes that describe old and new senses of a given word. These were generated using the XLM-R masked language model. For polysemous words, lexical substitutes depended on the meaning expressed in a particular context. For each target word, usages were sampled from both corpora, lemmatized and used to generate lexical substitutes. Next, two sets of vectors were built for old and new usages where each usage is represented by a vector of the probabilities of its substitutes. For Graded Change the Cosine APD between old and new vectors was computed, while for Binary Change a threshold was applied to this score. The authors also proposed three different approaches based on pairwise distances for the sense gain and loss subtasks.

\paragraph{Rombek} 
This system adapted ideas from the Word Sense Induction (WSI) task. Lexical substitutes were generated in the same way as with the BOS system (see above) and arranged in a matrix. Agglomerative clustering was then applied to each target word to obtain clusters with candidate senses. JSD was applied between clusters to obtain Graded Change estimates. Thresholding was applied to produce binary predictions.\footnote{This team did not submit a paper to the shared task.}

\subsection{Summary}
Most systems were based on three main components: (i) a semantic representation of words or word usages as vectors, (ii) an aggregation method over vectors, and (iii) a change measure. Type-based systems usually employed an additional alignment step over semantic representations. Also, the preprocessing of data was crucial for the performance of contextualized embeddings \citep{Laicher2021explaining}.

\paragraph{Preprocessing}
Some teams only used the tokenized version of the shared task dataset (CoToHiLi, UAlberta), while other teams only used the lemmatized version (UAlberta, BOS, HSE). One team varied the preprocessings with systems (UAlberta): lemmatization for type-based embeddings and tokenization, lemmatization and POS-tagging for the WSD system. Two teams did not use any sort of preprocessing (GlossReader, DeepMistake), while two teams used substitution with dynamic patterns (e.g. \textit{<mask> (y [target])}, \textit{[target] (por ejemplo <mask>)}) for their lexical substitution models (BOS, Rombek).

\paragraph{Semantic representations}
Most systems used token-based contextualized embeddings such as BERT (HSE) and XLM-R (DeepMistake, GlossReader, Rombek, UAlberta, BOS). Some teams further fine-tuned these embeddings on Language Modeling, WSD or WSI/WiC tasks. One team (DeepMistake) fine-tuned on the semantic proximity judgments from the published development data. Only three teams used type-based semantic representations including SGNS (UAlberta), CBOW (CoToHiLi) and Grammatical Profiling (HSE).

\paragraph{Vector aggregation}
Participating teams used different approaches to aggregate vectors into more abstract semantic representations. A common strategy was to model the COMPARE score by computing Average Pairwise Distances (APD) between vectors from different time periods (DeepMistake, GlossReader, UAlberta, BOS). This strategy has shown to perform well in various previous studies and shared tasks \citep{kutuzov-giulianelli-2020-uiouva,Laicher2021explaining,Kurtyigit2021discovery,Arefyev2021Deep}. Another strategy was to cluster the vectors (HSE, Rombek, DeepMistake). Clustering algorithms used are: Agglomerative Clustering (Rombek), K-means (HSE) and Correlation Clustering (DeepMistake). One system used a WSD system to assign cluster labels (UAlberta).

\paragraph{Change Measure}
For Graded Change most teams using contextualized embeddings directly relied on APD scores as described above. They used different distance measures such as: Cosine (BOS), Euclidean (UAlberta) and Manhattan (GlossReader) distances. One team averaged same-sense probabilities (DeepMistake). The teams relying on clustering mostly used the JSD to measure Graded Change (Rombek, DeepMistake). One team instead used cosine distance between cluster centroids (HSE). The teams relying on type-based representations used either Cosine (CoToHiLi, HSE) or Euclidean distance (UAlberta). For Binary Change most teams relied on thresholding the graded predictions (DeepMistake, GlossReader, Rombek, HSE, CoToHiLi, BOS). This strategy has shown high performance in several previous studies and shared tasks \citep{schlechtweg-etal-2020-semeval, kaiser-etal-2020-roots, Kurtyigit2021discovery}. Two teams using a clustering approach measured Binary Change by applying exactly the definition from the annotation process (DeepMistake) or a similar definition (UAlberta).

\section{Results}

\begin{table*}[htb]
    \centering
\begin{tabular}{|c|c|c|c|}
\multicolumn{2}{|c|}{Task} & Change graded & COMPARE \\ \hline 
\# & Team name & SPR & SPR \\ \hline
1 & GlossReader & \textbf{0.735 (1)} & 0.842 (1) \\
2 & DeepMistake & \textbf{0.702 (2)} & 0.829 (2) \\
3 & HSE & \textbf{0.553 (3)} & 0.558 (4) \\
4 & baseline1 & 0.543 (4) & 0.561 (3) \\
5 & baseline3 & 0.508 (5) & 0.459 (5) \\
6 & Rombek & 0.497 (6) & 0.456 (6) \\
7 & CoToHiLi & 0.282 (7) & -- \\
8 & baseline2 & 0.092 (8) & 0.088 (7) \\
9 & baseline5 & 0.064 (9) & -0.072 (8) \\
10 & BOS & -0.125 (10) & -0.129 (9) \\
\end{tabular}
    \caption{Summary of system performance in phase 1. Teams are ranked according to SPR score for the Graded Change subtask in decreasing order. The values corresponding to the three best systems are highlighted in bold type.}
    \label{tab:results_phase1}
\end{table*}

\begin{table*}[htb]
    \centering
\begin{tabular}{|c|c|c|c|c|c|c|}
\multicolumn{2}{|c|}{Task}  & \multicolumn{3}{c|}{Change binary} & Change graded & COMPARE   \\ \hline
\# & Team name & F1 & P & R & SPR & SPR   \\ \hline
1 & GlossReader & \textbf{0.716 (1)} & 0.615 (3) & 0.857 (3) & 0.735 (1) & 0.842 (1) \\
2 & UAlberta & \textbf{0.709 (2)} & 0.549 (7) & 1.000 (1) & -- & -- \\
3 & Rombek & \textbf{0.687 (3)} & 0.590 (4) & 0.821 (4) & 0.535 (5) & 0.546 (5) \\
4 & BOS &  0.658 (4) & 0.510 (8) & 0.929 (2) & 0.209 (8) & 0.163 (7) \\
5 & DeepMistake &  0.655 (5) & 0.633 (2) & 0.679 (6) & 0.676 (2) & 0.821 (2) \\
6 & CoToHiLi &  0.636 (6) & 0.553 (6) & 0.750 (5) & 0.282 (7) & --  \\
7 & baseline4 & 0.636 (6) & 0.467 (11) & 1.0 (1) & -- & -- \\
8 & HSE & 0.586 (7) & 0.567 (5) & 0.607 (7) & 0.553 (3) & 0.558 (4) \\
9 & baseline3  & 0.548 (8) & 0.500 (9) & 0.607 (7) & 0.373 (6) & 0.423 (6) \\
10 & baseline1 & 0.537 (9) & 0.846 (1) & 0.393 (9) & 0.543 (4) & 0.561 (3) \\
11 & baseline5 & 0.508 (10) & 0.484 (10) & 0.536 (8) & 0.064 (10)& -0.072 (9) \\
12 & baseline2 & 0.222 (11) & 0.500 (9) & 0.143 (10) & 0.092 (9) & 0.088 (8) \\

\end{tabular}
    \caption{Summary of the results of Phase 2 for substasks Graded Change, COMPARE and Binary Change. Teams are ranked according to F1 score for subtask Change binary in decreasing order. The values corresponding to the three best systems are highlighted in bold type.}
    \label{tab:results_phase2_part1}
\end{table*}

\begin{table*}[htb]
    \centering
\begin{tabular}{|c|c|c|c|c|c|c|c|}
\multicolumn{2}{|c|}{Task} &  \multicolumn{3}{c|}{Sense gain} & \multicolumn{3}{c|}{Sense loss} \\ \hline
\# & Team name & F1 & P & R & F1 & P & R \\ \hline
1 & GlossReader & \textbf{0.511 (3)} & 0.333 (5) & 0.929 (2) & \textbf{0.688 (1)} & 0.564 (2) & 0.880 (2) \\
2 & DeepMistake & \textbf{0.591 (1)} & 0.433 (1) & 0.929 (2) & 0.582 (5) & 0.533 (3) & 0.640 (4) \\
3 & HSE & 0.250 (8) & 0.192 (9) & 0.357 (5) & 0.364 (7) & 0.421 (5) & 0.320 (5) \\
4 & baseline1 & -- & -- & -- & -- & -- & -- \\
5 & Rombek & 0.50 (4) & 0.409 (2) & 0.643 (4) & \textbf{0.681 (2)} & 0.727 (1) & 0.640 (4) \\
6 & baseline3 & -- & -- & -- & -- & -- & -- \\
7 & BOS & \textbf{0.520 (2)} & 0.361 (4) & 0.929 (2) & \textbf{0.610 (3)} & 0.529 (4) & 0.720 (3) \\
8 & baseline2 & 0.211 (9) & 0.400 (3) & 0.143 (6) & 0 (8) & 0 (8) & 0 (7) \\
9 & UAlberta & 0 (10) & 0 (10) & 0 (7) & 0 (8) & 0 (8) & 0 (7) \\
10 & CoToHiLi & 0.462 (5) & 0.316 (6) & 0.857 (3) & 0 (8) & 0 (8) & 0 (7) \\
11 & baseline4 & 0.378 (6)& 0.23 (8)& 1.0 (1)& 0.588 (4)& 0.416 (6)& 1.0 (1)\\
12 & baseline5 & 0.333 (7)& 0.313 (7)& 0.357 (5)& 0.367 (6)& 0.375 (7)& 0.36 (6)\\
\end{tabular}
    \caption{Summary of the results of Phase 2 for subtasks Sense loss and Sense gain. The values corresponding to the three best systems are highlighted in bold type.}
    \label{tab:results_phase2_part2}
\end{table*}

The results shown in Tables \ref{tab:results_phase1}, \ref{tab:results_phase2_part1} and \ref{tab:results_phase2_part2} correspond to the best submissions per subtask.\footnote{In the case of HSE who used two different systems, the displayed results correspond to the token-based system. \todo[inline]{describe after the extraction of the values}} 
\todo[inline]{extract the best submissions for COMPARE, sense gain and sense loss}

\paragraph{Graded Change Discovery} As shown in Table \ref{tab:results_phase1}, \textbf{GlossReader} and \textbf{DeepMistake} obtained first and second place in the main task of evaluation phase 1, while \textbf{HSE} came third.\footnote{Since not not all users reported a team name on Codalab, some leaderboard entries are filled with usernames.} These were the only teams that managed to outperform baseline1 (SGNS+OP+CD) and baseline3 (Grammatical Profiles). The three winning systems were based on fine-tuned versions of contextualized embeddings with average vector aggregation (GlossReader, DeepMistake) or clustering (HSE). Interestingly, the top two systems did not model the JSD between cluster distributions (as done on the annotation to derive gold scores), but instead model the COMPARE score (with APD). We discuss this observation further in Subsection \ref{sub:discussion}.

\paragraph{COMPARE Discovery} GlossReader and DeepMistake also reached the first and second place on the COMPARE task in evaluation phase 1. This is not surprising, because they actually modeled the COMPARE score with APD. Consequently, also the correlation was considerably higher than with Graded Change (e.g. $\rho=0.842$ vs. $0.735$). Baseline1 took the third place. 

\paragraph{Binary Change Detection} For Phase 2 (Tables~\ref{tab:results_phase2_part1} and~\ref{tab:results_phase2_part2}), again \textbf{GlossReader} performed best, this time followed by \textbf{UAlberta} and \textbf{Rombek}. Interestingly, with the exception of GlossReader the systems used in Phase 1 did not obtain a good performance in Phase 2. However, participants managed to outperform all baselines with the exception of HSE not outperforming baseline4 (minority class). Two out of the winning systems used thresholding (GlossReader, Rombek), i.e., they modeled the COMPARE score or the JSD and then thresholded these scores to obtain Binary Change predictions. From these teams only UAlberta inferred sense clusters. Hence, here we saw again what we saw for phase 1: the top-performing teams were often not modeling the annotation procedure.

\paragraph{Sense Gain/Loss Detection} The top performance for sense gain ($\operatorname{F1}=0.591$) was clearly lower than for Binary Change, while for loss the top performance ($\operatorname{F1}=0.688$) approaches the one for Binary Change. The best results for sense gain were obtained by \textbf{DeepMistake}, followed by \textbf{BOS} and \textbf{GlossReader}. In the sense loss subtask, \textbf{GlossReader} obtained the best performance, followed by \textbf{Rombek} and \textbf{BOS}. GlossReader and DeepMistake submitted the same results to both subtasks as for Binary Change Detection implicitly assuming that gain and loss always occur together. In this way, they mostly outperformed Rombek and BOS who tried a more principled approach.

\paragraph{Graded Change/COMPARE Detection} The top performance for these tasks was the same in evaluation phase 1 and 2 ($\rho =0.735$ and $0.842$). Some teams had the same results in both phases (GlossReader, HSE, CoToHiLi) and thus likely submitted the same predictions. Two teams improved their results (Rombek, BOS), while one team had lower results (DeepMistake). We are unsure about the impact of the published target words and their usages on these results, as teams did not consistently report whether they used this information in phase 2.

\subsection{Discussion}\label{sub:discussion}

\begin{figure*}[htb]
\begin{center}
     \includegraphics[width=0.7\linewidth]{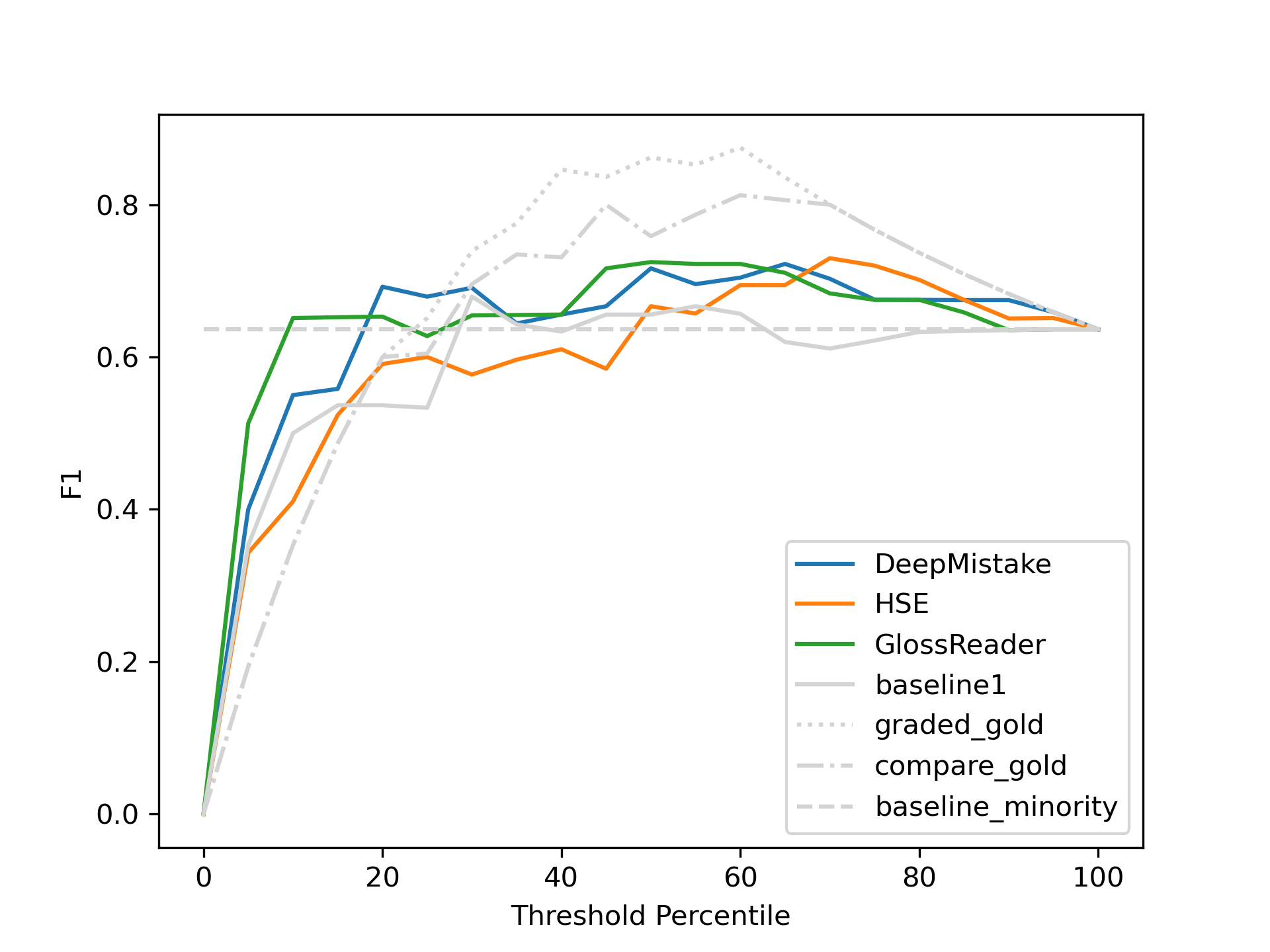} \caption{F1 scores over binarization thresholds based on percentiles on submitted Graded Change predictions for top four teams in evaluation phase 1.}\label{fig:thresholds}
\end{center}
\end{figure*}

The Graded Change Discovery subtask was solved with a rather high performance by the winning team ($\rho=0.735$). This is comparable to the top performance in SemEval ($\rho=0.725$ for DE) obtained with type-based embeddings. The COMPARE Discovery subtask was solved with even higher performance ($\rho=0.842$). This is comparable to the top performance in RuShiftEval ($\rho=0.822$). However, the results in our shared task were obtained under harder conditions, i.e., for a large number of uncleaned target words (Discovery).\footnote{We assume that the performance of participating systems obtained on the hidden target words generalizes roughly to the full set of public target words as the sample was taken largely random.} This suggests that, as far as Graded Change is concerned, LSCD systems are applicable to solve real-world problems and may be useful in historical semantics or lexicography. However, the more relevant task for these fields is Binary Change Detection/Discovery \citep{schlechtweg-walde-2020}. The results for Binary Change Detection were lower ($\operatorname{F1}=0.716$), but still clearly higher than the best baseline ($0.636$). Results in SemEval were mixed, but mostly not higher than $\operatorname{F1}=0.7$ (DE), while results in DIACR-Ita were high with an accuracy of $0.94$, which was, however, obtained with a different metric and on a very small and strongly preselected set of target words. A future challenge will thus be to improve performance on the binary task.

Our shared task was clearly dominated by token-based systems. Out of seven participants only two used a (standalone) type-based system which also performed much worse than the winning teams (CoToHiLi, HSE).\footnote{The result reported by the HSE team in the leaderboard corresponds to the first method described in Section~\ref{sec:partsystems}.} Also, our type-based baseline1 was clearly outperformed by a number of token-based systems (three in phase 1 and six in phase 2). This confirms the tendency observed in RuShiftEval where token-based systems outperformed type-based ones on LSCD. Before that, in SemEval and DIACR-Ita the type-based systems had dominated. Potential reasons for this switch are the understanding of biases in contextualized embeddings \citep{Laicher2021explaining}, their optimization through fine-tuning \citep{Arefyev2021Deep, Arefyev2021Interpretable} and the optimization of vector aggregation methods \citep{kutuzov-giulianelli-2020-uiouva, Laicher2021explaining, Arefyev2021Deep}. 

In our task, we saw clustering methods amongst the best-performing systems (HSE, UAlberta) for the first time. This is an important development, because the current top-performing system (GlossReader), as well as many other systems not relying on clustering, did not model the target word annotation procedure (cf. Subsection \ref{sub:annotation}). Instead, it exploited correlations between the COMPARE score and JSD as well as Binary Change. These scores are known to correlate strongly in current LSCD datasets \citep{Schlechtweg2022measurement}, including ours. The correlation between gold (negated) COMPARE and JSD scores in our dataset is $0.92$, while it is $0.69$ for gold (negated) COMPARE and Binary Change. This means that modeling the COMPARE score is a good predictor for Graded as well as Binary Change. However, this also means that, the current best-performing systems have a clear upper bound on their potential to solve LSCD tasks (where this upper bound is higher for Graded than for Binary Change). Hence, if we want to break through this upper bound in the future, we need to develop or improve other system types possibly relying on clustering to model the annotation procedure.\footnote{\citet{deepmistake-lcsdiscovery} had promising results with applying the clustering framework used in the annotated data and semantic proximity graphs derived from fine-tuned contextualized embeddings.} 

In order to see how far the current approach of thresholding COMPARE/JSD/graded scores carries, we compared performance of the top three systems in evaluation phase 1 across binarization thresholds in Figure \ref{fig:thresholds}. As we can see, the three systems had a similar maximum performance of roughly $\operatorname{F1}=0.72$ around a binarization threshold of $50-70$ \%.\footnote{Interestingly, HSE here obtained maximum performance amongst all systems ($0.73$), much higher than their submission in evaluation phase 2. A similar observation holds for our baseline1. This shows how crucial threshold selection is in this approach.} At $100$ \% they all converged to the minority class baseline (all target words labeled as $1$). The upper bound on this approach was given by the maximum performance of the gold JSD (graded\_gold) and the gold COMPARE score (compare\_gold). These upper bounds were $0.88$ and $0.81$ respectively. This means that perfectly modeling the COMPARE or even the JSD score can reach high but never perfect performance on Binary Change.

\section{Conclusion}\label{sec:conclusions}

We conducted the first shared task on semantic change discovery and detection in Spanish. We manually annotated $100$ Spanish words for semantic change between two corpora, an old one covering the period between 1810 and 1906, and a modern one covering the years between 1994 and 2020. The discovery part of our task imposed several computational challenges for participants, as it required calculating semantic change scores for all words in the vocabulary.

We received predictions from six teams in phase 1 and seven teams in phase 2. Participants applied systems using static and contextualized word embeddings in combination with various fine-tuning procedures, vector aggregation methods and change measures. Graded Change Discovery was solved with high performance while Binary Change Detection still remains far from being solved. The most successful method winning both main tasks is a system fine-tuning contextualized multilingual XML-R embeddings on WSD data, aggregating vectors into cross-corpus pairs and measuring change as the average of their distances, or a binarization of these values. However, we showed that this approach has a clear upper bound which will not allow to solve the tasks completely reliably in the future. Another interesting result from our task was that clustering approaches are amongst the winning teams for the first time.

We hope that this shared task will help pave the way for future research in the discovery and detection of semantic lexical changes for the Spanish language, and that our data can be used in the future for the proposal of novel ideas and techniques.

\section{Acknowledgements}
This work was supported by ANID FONDECYT grant 11200290, U-Inicia VID Project UI-004/20, ANID -Millennium Science Initiative Program - Code ICN17\_002, the National Center for Artificial Intelligence CENIA FB210017, Basal ANID, and SemRel Group (DFG Grants SCHU 2580/1 and SCHU 2580/2). Dominik Schlechtweg has been funded by
the project `Towards Computational Lexical Semantic Change Detection' supported by the Swedish Research Council (2019–2022; contract 2018-01184) and by the research program `Change is Key!' supported by Riksbankens Jubileumsfond (under reference number M21-0021).

% Entries for the entire Anthology, followed by custom entries
\bibliography{anthology,bibliography-self,Bibliography-general}
\bibliographystyle{acl_natbib}

\appendix

\section*{Appendix}
\label{sec:appendix}

\section{Lemmatization}\label{lemmatizer-related-issues}
\label{sec:lemma}
Manual inspection showed that spaCy sometimes yielded erroneous lemmatization. This happened more frequently for sentences in the old corpus and for tokens at the beginning of sentences as shown in the example below:

\begin{center}
\fbox
{
\begin{minipage}{0.45\textwidth}
\small
\textbf{Example}:\\
"Decidióse ésta por Teresa la expósita, y así se vio a la vagamunda
tomar bajo su amparo a la pobre desheredada como ella."

\textbf{Lemmatization}:\\
Decidióse este por Teresa el expósita , y así él ver a el vagamunda
tomar bajo su amparo a el pobre desheredado como él .

\end{minipage}
}
\end{center}
As can be seen, the lemma of the word \emph{Decidióse} was not found,
nor was the word converted to lowercase.
SpaCy version 3.1.1 with es\_core\_news\_md (3.1.0) was used.

\section{Target indices of annotated usages}
In the first version of the extracted word usages which were uploaded to the DURel interface for annotation there were frequent errors for the target word indices. As a result, the wrong target words were marked in these usages. However, annotators were instructed to search for the correct target words and to judge these instead. We corrected the indices for the data provided to participants during the shared task. However, we later noticed that some indices included punctuation immediately following the target word as shown below:
\begin{center}
\fbox
{
\begin{minipage}{0.45\textwidth}
\small
\textbf{Example}\\
lemma: sexo \\
context:  136. Los apellidos de familia no varían de terminación para los diferentes \textbf{sexos;} y así se dice «don Pablo Herrera», «doña Juana Hurtado», «doña Isabel Donoso». 137 (b). \\
indexes\_target\_token: 75:81

\end{minipage}
}
\end{center}
After the shared task we uploaded a data version with corrected indices.

\end{document}